\documentclass{article} 
\usepackage{iclr2017_conference,times}
\usepackage{hyperref}
\usepackage{url}

\usepackage{hyperref}       
\usepackage{url}            
\usepackage{booktabs}       
\usepackage{amsfonts}       
\usepackage{nicefrac}       
\usepackage{microtype}      

\usepackage{amsmath}
\usepackage{algorithm}
\usepackage{algorithmic}
\usepackage{bm}
\usepackage{multirow}
\usepackage{graphicx}
\usepackage{wrapfig}

\DeclareMathOperator*{\Exp}{\mathbb{E}}

\title{An Actor-Critic Algorithm\\
       for Sequence Prediction}

\author{Dzmitry Bahdanau~  ~\textbf{Philemon Brakel} \\ 
        \textbf{Kelvin Xu}~ ~\textbf{Anirudh Goyal} \\ 
        Universit\'e de Montr\'eal \\
        \And
        Ryan Lowe~ ~\textbf{Joelle Pineau}\thanks{CIFAR Senior Fellow} \\
        McGill University \\
        \AND
        Aaron Courville\thanks{CIFAR Fellow}\\
        Universit\'e de Montr\'eal \\
        \And
        \hspace{52mm}Yoshua Bengio\footnotemark[1] \\
        \hspace{52mm}Universit\'e de Montr\'eal
}

\iclrfinalcopy 

\begin{document}

\maketitle

\begin{abstract}
    We present an approach to training neural networks to generate sequences using actor-critic methods from reinforcement learning (RL).
    Current log-likelihood training methods are limited by the discrepancy between their training and testing modes, 
    as models must generate tokens conditioned on their previous guesses rather
    than the ground-truth tokens. 
    We address this problem by introducing a \textit{critic} network that is trained to predict the value of an output token, given the policy of an \textit{actor} network.
    This results in a training procedure that is much closer to the test phase, and allows us to directly optimize for a task-specific score such as BLEU.
    Crucially, since we leverage these techniques in the supervised learning
    setting rather than the traditional RL setting, we condition the critic
    network on the ground-truth output.
    We show that our method leads to improved performance on both a synthetic task, and for German-English machine translation.
 Our analysis paves the way for such methods to be applied in natural 
 language generation tasks, such as machine translation, caption generation, and dialogue modelling.
\end{abstract}

\section{Introduction}

In many important applications of machine learning, the task is to develop a
system that produces a sequence of discrete tokens given an input. Recent work
has shown that recurrent neural networks (RNNs) can deliver excellent
performance in many such tasks when trained to predict the next output token
given the input and previous tokens. This approach has been applied successfully in machine translation
\citep{sutskever2014sequence,bahdanau2015neural}, caption generation
\citep{kiros2014unifying,donahue2015long,vinyals2015show,xu2015show,karpathy2015deep}, and speech recognition 
\citep{chorowski2015attention,chan2015listen}. 

The standard way to train RNNs to generate sequences is to
maximize the log-likelihood of the ``correct'' token given a history of the
previous ``correct'' ones, an approach often called \textit{teacher forcing}.
At evaluation time, the output sequence is often produced by an approximate search for 
the most likely candidate according to the learned distribution. During this search,
the model is conditioned on its own guesses, which may be incorrect and thus lead to a compounding of errors~\citep{bengio2015scheduled}.
This can become especially problematic for longer sequences.
Due to this discrepancy between training and testing conditions, it has been shown that maximum likelihood training can be suboptimal \citep{bengio2015scheduled,ranzato2015sequence}. 
In these works, the authors argue that the network should be trained to continue
generating correctly given the outputs already produced by the model, rather
than the \emph{ground-truth} reference outputs from the data. This gives rise to the challenging problem of
determining the target for the next network output.
\citet{bengio2015scheduled} use the
token $k$ from the ground-truth answer as the target for the network at step
$k$, whereas \citet{ranzato2015sequence} rely on the REINFORCE
algorithm \citep{williams1992simple} to decide whether or not the tokens from a
sampled prediction lead to a high task-specific score, such as BLEU 
\citep{papineni2002bleu} or ROUGE \citep{lin2003automatic}.

In this work, we propose and study an alternative procedure for training
sequence prediction networks that aims to directly improve their test time
metrics (which are typically not the log-likelihood).
In particular, we train an additional network called 
the \textit{critic} to output the \textit{value} of each
token, which we define as the expected task-specific score that the network
will receive if it outputs the token and continues to sample outputs according to
its probability distribution. Furthermore, we show how the predicted values can be used
to train the main sequence prediction network, which we refer to as the \textit{actor}.
The theoretical foundation of our method is that, under the assumption that the
critic computes exact values, the expression that we use to train the actor is an unbiased
estimate of the gradient of the expected task-specific score. 

Our approach draws inspiration and borrows the terminology from the field of
reinforcement learning (RL) \citep{sutton1998introduction}, in particular from
the actor-critic approach \citep{sutton1984temporal,sutton1999policy,barto1983neuronlike}.  RL
studies the problem of acting efficiently based
only on weak supervision in the form of a reward given for some of the agent's actions.
In our case, the reward is analogous to the task-specific score associated with a prediction.
However, the tasks we consider are those of \textit{supervised learning}, and
we make use of this crucial difference by allowing the critic to \textit{use the
ground-truth answer as an input}.
In other words, the critic has access to a sequence of expert actions that are
known to lead to high (or even optimal) returns. 
To train the critic, we adapt the temporal
difference methods from the RL literature \citep{sutton1988learning} to our setup.
While RL methods with non-linear function approximators are not new~\citep{tesauro1994td,miller1995neural}, 
they have recently surged in popularity, giving rise to the field of `deep RL'
\citep{mnih2015human}. We show that some of the techniques recently developed in deep RL, such as having a \textit{target network}, may also be beneficial for sequence prediction.

The contributions of the paper can be summarized as follows: 1) we describe how
RL methodology like the actor-critic approach can be applied to
supervised learning problems with structured outputs; and 2) we investigate the
performance and behavior of the new method on both a synthetic task and a
real-world task of machine translation, demonstrating the improvements over
maximum-likelihood and REINFORCE brought by the actor-critic training.

\section{Background}
We consider the problem of learning to produce an output sequence $Y=(y_1,
\ldots, y_T)$, $y_t \in \mathcal{A}$ given an input $X$, where $\mathcal{A}$ is
the alphabet of output tokens. We will often use notation $Y_{f \ldots l}$ to
refer to subsequences of the form $(y_f, \ldots, y_l)$.
Two sets of input-output pairs $(X, Y)$ are
assumed to be available for both training and testing. The
trained predictor $h$ is evaluated by computing the average task-specific score
$R(\hat{Y}, Y)$ on the test set, where $\hat{Y}=h(X)$ is the prediction.
 To
simplify the formulas we always use $T$ to denote the length of an output sequence, ignoring the fact that the output sequences may have different length.

\paragraph{Recurrent neural networks}

A recurrent neural network (RNN) produces a sequence of state vectors $(s_1,
\ldots, s_T)$ given a sequence of input vectors $(e_1, \ldots, e_T)$ by
starting from an initial $s_0$ state and applying $T$ times the transition
function $f$: $s_{t} = f(s_{t-1}, e_t)$. Popular choices for the mapping $f$ are the Long Short-Term Memory \citep{hochreiter1997long} and the Gated
Recurrent Units \citep{cho2014learning}, the latter of which we use for our models. 

To build a probabilistic model for sequence generation with an RNN,
one adds a stochastic output layer $g$ (typically a softmax for discrete outputs)
that generates outputs $y_t \in \mathcal{A}$ 
and can feed these outputs back by replacing them with their embedding $e(y_t)$:
\begin{align}
    y_t \sim g(s_{t-1}) \\
    s_t = f(s_{t - 1}, e(y_t)).
\end{align}    

Thus, the RNN defines a probability distribution $p(y_t|y_1, \ldots, y_{t-1})$
of the next output token $y_t$ given the previous tokens $(y_1, \ldots, y_{t-1})$.
Upon adding a special end-of-sequence token $\emptyset$ to the alphabet
$\mathcal{A}$, the RNN can define the distribution $p(Y)$ over all possible
sequences as $p(Y) = p(y_1) p(y_2 | y_1) \ldots p(y_T|y_1, \ldots, y_{T - 1})
p(\emptyset | y_1, \ldots, y_T)$.

\paragraph{RNNs for sequence prediction}
To use RNNs for sequence prediction, they must be augmented 
to generate $Y$ conditioned on an input $X$. The simplest way to do this
is to start with an initial state $s_0=s_0(X)$ 
\citep{sutskever2014sequence,cho2014learning}. Alternatively, one can 
encode $X$ as a variable-length sequence of vectors $(h_1, \ldots, h_L)$ and condition
the RNN on this sequence using an attention mechanism.
In our models, the sequence of vectors is produced by either a bidirectional RNN
\citep{schuster1997bidirectional} or a convolutional encoder \citep{rush2015neural}.

We use a \emph{soft} attention mechanism~\citep{bahdanau2015neural} that computes a weighted sum
of a sequence of vectors. The attention weights
determine the relative importance of each vector.
More formally, we consider the following equations for RNNs with attention:
\begin{align}
    y_t \sim g(s_{t-1}, c_{t-1}) \\ 
    s_t = f(s_{t-1}, c_{t-1}, e(y_t)) \\
    \alpha_t = \beta(s_t, (h_{1},\ldots,h_{L})) \\
    c_t = \sum\limits_{j=1}^L \alpha_{t,j} h_j
\end{align}   
where $\beta$ is the attention mechanism that produces the attention weights
$\alpha_t$ and $c_t$ is the context vector, or `glimpse', for time step $t$.
The attention weights are computed by an MLP that takes as input the current RNN
state and each individual vector to focus on. The weights are typically (as in
our work) constrained to be positive and sum to 1 by using the softmax function.

A conditioned RNN can be trained for sequence prediction by gradient ascent on
the log-likelihood $\log p(Y|X)$ for the input-output pairs $(X, Y)$ from the
training set. To produce a prediction $\hat{Y}$ for a test input sequence $X$, an approximate
beam search for the maximum of $p(\cdot|X)$ is usually conducted. During this
search the probabilities $p(\cdot|\hat{y}_1, \ldots, \hat{y}_{t-1})$ are considered,
where the previous tokens $\hat{y}_1, \ldots, \hat{y}_{t-1}$ comprise a candidate
beginning of the prediction $\hat{Y}$. 


\begin{algorithm}
    \caption{Actor-Critic Training for Sequence Prediction}
    \label{algo:algo1}
    \begin{algorithmic}[1]
    \REQUIRE 
        A critic $\hat{Q}(a;\hat{Y}_{1 \ldots t}, Y)$
        and an actor $p(a|\hat{Y}_{1 \ldots t}, X)$ with
        weights $\phi$ and $\theta$ respectively.
    \STATE Initialize delayed actor $p^{'}$ 
    and target critic $\hat{Q}'$ with same weights: $\theta' = \theta$, $\phi' = \phi$.
    \WHILE{Not Converged}
    \STATE Receive a random example $(X, Y)$.
    \STATE Generate a sequence of actions $\hat{Y}$ from $p^{'}$.
    \STATE Compute targets for the critic
    \begin{equation*}
        q_t = r_t(\hat{y}_t;\hat{Y}_{1 \ldots t - 1}, Y)
        + \sum\limits_{a \in \mathcal{A}}
        p^{'}(a|\hat{Y}_{1 \ldots t}, X)
        \hat{Q}'(a;\hat{Y}_{1 \ldots t}, Y)
    \end{equation*}       
    \STATE Update the critic weights $\phi$ using the gradient
    \begin{align*}
    \frac{d}{d\phi}\left(
        \sum_{t=1}^{T} 
        \left(\hat{Q}(\hat{y}_t;\hat{Y}_{1 \ldots t - 1}, Y) - q_t \right)^2
        + \lambda_C C_t
    \right) \\
    \textrm{where } C_t = \sum_{a}\left( 
        \hat{Q}(a; \hat{Y}_{1 \ldots t-1}) - 
        \frac{1}{|\mathcal{A}|}\sum_{b}\hat{Q}(b; \hat{Y}_{1 \ldots t-1}) 
    \right)^2
    \end{align*}
    \STATE Update actor weights $\theta$ using the following gradient estimate
    \begin{align*}
        \widehat{\frac{dV(X,Y)}{d\theta}} &=
        \sum\limits_{t=1}^T
        \sum\limits_{a \in \mathcal{A}}
        \frac{d p(a|\hat{Y}_{1 \ldots t - 1}, X)}{d \theta}
         \hat{Q}(a;\hat{Y}_{1 \ldots t - 1}, Y)
\\
         &+ \lambda_{LL} \sum\limits_{t=1}^T 
        \frac{d p(y_t|Y_{1 \ldots t - 1}, X)}{d \theta}
    \end{align*}
    \STATE Update delayed actor and target critic, with constants 
    $\gamma_{\theta} \ll 1$, $\gamma_{\phi} \ll 1$
        \begin{align*}
            \theta' = \gamma_{\theta} \theta + (1 - \gamma_{\theta}) \theta',\, 
            \phi' = \gamma_{\phi} \phi + (1 - \gamma_{\phi}) \phi'
        \end{align*}
    \ENDWHILE
    \end{algorithmic}
\end{algorithm}

\begin{algorithm}
    \caption{Complete Actor-Critic Algorithm for Sequence Prediction}
    \label{algo:algo2}
    \begin{algorithmic}[1]
    \STATE 
        Initialize critic $\hat{Q}(a;\hat{Y}_{1 \ldots t}, Y)$
        and actor $p(a|\hat{Y}_{1 \ldots t}, X)$ with
        random weights $\phi$ and $\theta$ respectively.
    \STATE 
        Pre-train the actor to predict $y_{t+1}$ given 
        $Y_{1 \ldots t}$ by maximizing 
        $\log p(y_{t+1}|Y_{1 \ldots t}, X)$.
    \STATE 
        Pre-train the critic to estimate $Q$ by running Algorithm \ref{algo:algo1}
        with fixed actor.
    \STATE
        Run Algorithm \ref{algo:algo1}.
    \end{algorithmic}
\end{algorithm}

\paragraph{Value functions}

We view the conditioned RNN as a stochastic \textit{policy}
that generates \textit{actions}
and receives the task score (e.g., BLEU score)  as the \textit{return}.  
We furthermore consider the case when
the return $R$ is partially received at the intermediate steps 
in the form of \textit{rewards} $r_t$:  $R(\hat{Y}, Y) =
\sum_{t=1}^T r_t(\hat{y}_t;\hat{Y}_{1 \ldots t-1}, Y)$.  This
is more general than the case of receiving the full return at the end of the sequence, 
as we can simply define all rewards other than $r_T$ to be zero.
Receiving intermediate rewards may ease the learning for the critic, and we use
\emph{reward shaping} as explained in Section \ref{sec:actor-critic}. 
Given the policy, possible actions and reward function, the \emph{value} represents the
expected future return as a function of the
current state of the system, which in our case is uniquely defined by the
sequence of actions taken so far, $\hat{Y}_{1 \ldots t-1}$.

We define the value of an unfinished
prediction $\hat{Y}_{1 \ldots t}$ as follows:
\begin{equation*}
    V(\hat{Y}_{1 \ldots t}; X, Y) = 
    \Exp_{
        \hat{Y}_{t+1 \ldots T}
        \sim p(.| \hat{Y}_{1 \ldots t}, X)
    } \sum\limits_{\tau=t+1}^T r_{\tau}(\hat{y}_{\tau}; \hat{Y}_{1 \ldots \tau - 1}, Y).
\end{equation*}    
We define the value of a candidate next token $a$ for an unfinished prediction 
$\hat{Y}_{1 \ldots t - 1}$ as the expected future return after generating token $a$: 
\begin{align*}
    Q(a; \hat{Y}_{1 \ldots t - 1}, X, Y) = 
    \Exp_{
        \hat{Y}_{t+1 \ldots T}
        \sim p(.|\hat{Y}_{1 \ldots t - 1} a,  X)
    } 
    \left(
    r_{t}(a; \hat{Y}_{1 \ldots t-1}, Y) + 
    \sum\limits_{\tau=t+1}^{T} 
    r_{\tau}(\hat{y}_{\tau}; \hat{Y}_{1 \ldots t-1} a \hat{Y}_{t+1 \ldots \tau}, Y)
    \right).        
\end{align*}
We will refer to the candidate next tokens as \textit{actions}.
For notational simplicity, we henceforth drop $X$ and $Y$ from the signature
of $p$, $V$, $Q$, $R$ and $r_t$, assuming it is clear from the context which of $X$ and $Y$ is
meant.  We will also use $V$ without arguments for the expected reward of
a random prediction.

\section{Actor-Critic for Sequence Prediction}

\label{sec:actor-critic}

Let $\theta$ be the parameters of the conditioned RNN, which we will also refer to as the \textit{actor}.
Our training algorithm is based on the following way of rewriting the gradient 
of the expected return $\frac{dV}{d\theta}$:
 \begin{align}
    \frac{dV}{d\theta} = 
    \Exp_{\hat{Y} \sim p(\hat{Y}|X)} 
    \sum\limits_{t=1}^T
    \sum\limits_{a \in \mathcal{A}}
    \frac{dp(a|\hat{Y}_{1 \ldots t - 1})}{d\theta}
    Q(a;\hat{Y}_{1 \ldots t - 1})
    \label{eq:ac}.
\end{align}
This equality is known in RL under the names policy gradient theorem \citep{sutton1999policy}
and stochastic actor-critic~\citep{sutton1984temporal}.
\footnote{We also provide a simple self-contained proof of Equation \eqref{eq:ac} in Supplementary Material.}
Note that we use the probability rather than the log probability in this formula 
(which is more typical in RL applications) as we are summing over actions rather than taking an expectation.  
Intuitively, this equality corresponds to increasing the probability of actions 
that give high values, and decreasing the probability of actions that give low values. 
Since this gradient expression is an expectation,
it is trivial to build an unbiased estimate for it:
\begin{align}
    \widehat{\frac{dV}{d\theta}} = 
    \sum\limits_{k=1}^M
    \sum\limits_{t=1}^T
    \sum\limits_{a \in \mathcal{A}}
    \frac{dp(a|\hat{Y}^k_{1 \ldots t - 1})}{d\theta}
    Q(a;\hat{Y}^k_{1 \ldots t - 1})
    \label{eq:actor_critic}
\end{align}
where $\hat{Y}^k$ are $M$ random samples from $p(\hat{Y})$. By replacing $Q$
with a parameteric estimate $\hat{Q}$ one can obtain a biased estimate with
relatively low variance. The parameteric estimate $\hat{Q}$ is
called the \textit{critic}.

The above formula is similar in spirit to the REINFORCE learning rule that \citet{ranzato2015sequence} use in the same context:
\begin{align}
    \widehat{\frac{dV}{d\theta}} = 
    \sum\limits_{k=1}^M
    \sum\limits_{t=1}^T
    \frac{ d\log p(\hat{y}^k_t|\hat{Y}^k_{1 \ldots t - 1}) }{d\theta}
    \left[
    \sum_{\tau=t}^T r_{\tau}(\hat{y}^k_{\tau}; \hat{Y}^k_{1 \ldots \tau - 1})
    - b_t(X)
    \right],
\end{align}
where the scalar $b_t(X)$ is called \textit{baseline} or \textit{control variate}.
The difference is that in REINFORCE the inner
sum over all actions is replaced by its 1-sample estimate, namely
$\frac{d \log p(\hat{y}_t|\hat{Y}_{1 \ldots t - 1})}{d\theta}
Q(\hat{y}_t;\hat{Y}_{1 \ldots t - 1})$, where the log probability 
$\frac{d \log p(\hat{y}_t|...)}{d\theta}=
\frac{1}{p(\hat{y}_t|...)}
\frac{d p(\hat{y}_t|...)}{d\theta}$ is introduced
to correct for the sampling of $\hat{y_t}$.
Furthermore, instead of the value $Q(\hat{y}_t;\hat{Y}_{1 \ldots t - 1})$, REINFORCE uses
the cumulative reward  
$\sum_{\tau=t}^T r_{\tau}(\hat{y}_{\tau}; \hat{Y}_{1 \ldots \tau - 1})$
following the action $\hat{y}_t$, which again can be seen as a 1-sample estimate of $Q$.
Due to these simplifications and the potential high variance in the cumulative reward, the REINFORCE
gradient estimator has very high variance. In order to improve upon it, we
consider the actor-critic estimate from Equation \ref{eq:actor_critic}, which has a lower variance 
at the cost of significant bias, since the critic is not perfect and trained simultaneously
with the actor. The success depends on our ability to control the bias
by designing the critic network and using an appropriate training criterion for it.

\begin{figure}
    \centering
    \includegraphics[scale=.7]{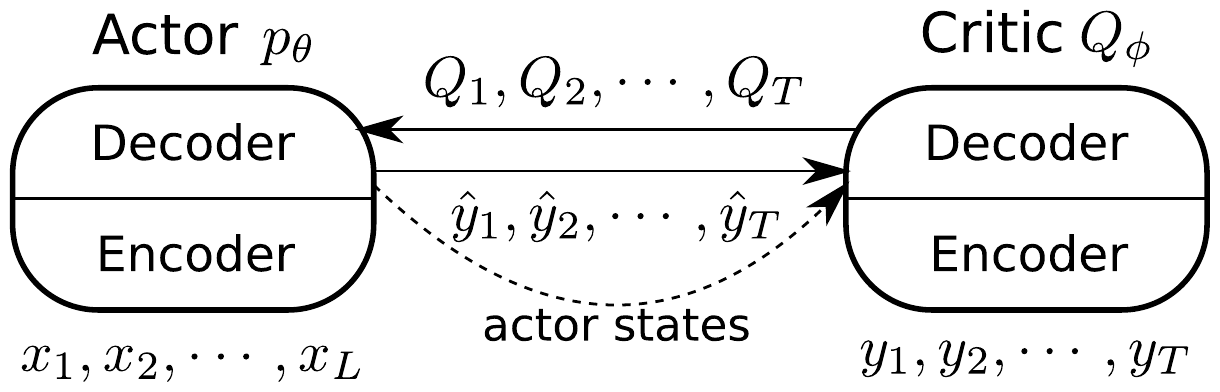}
    \caption{Both the actor and the critic are encoder-decoder networks. The actor receives an input sequence $X$ and produces samples
        $\hat{Y}$ which are evaluated by the critic. The critic takes in the ground-truth 
        sequence $Y$ as input to the encoder, 
        and takes the input summary (calculated using an attention mechanism) and the 
        actor's prediction $\hat{y}_t$ as input at time step $t$ of the decoder.  
    The values $Q_1,Q_2,\cdots,Q_T$ computed by the critic are used to approximate the gradient of the
    expected returns with respect to the parameters of the actor. This gradient
    is used to train the actor to optimize these expected task specific returns
    (e.g., BLEU score). The critic may also receive the hidden state activations of the actor as input.}
\label{fig:arch}
\end{figure}

To implement the critic, we propose to use a separate RNN parameterized by $\phi$.
The critic RNN is run
 in parallel with the actor, consumes the tokens $\hat{y}_t$
that the actor outputs and
 produces the estimates $\hat{Q}(a;\hat{Y}_{1
\ldots t})$ for all $a \in \mathcal{A}$. 
 A key difference between the critic
and the actor is that the
 correct answer $Y$ is given to the critic as an
input, similarly to how the actor
 is conditioned on $X$. Indeed, the return
$R(\hat{Y}, Y)$ is a deterministic
 function of $Y$, and we argue that using
$Y$ to compute $\hat{Q}$ should be of
 great help.  We can do this because the
values are only required during training and we do not use the critic at test
time.  We also experimented with providing the actor states $s_t$ as additional
 inputs to the critic.
 See Figure \ref{fig:arch} for a visual
representation of our actor-critic architecture.

\paragraph{Temporal-difference learning}
A crucial component of our approach is \emph{policy evaluation}, that is the training of the critic
to produce useful estimates of $\hat{Q}$. With a na{\"i}ve Monte-Carlo method, one could use the future
return $\sum_{\tau=t}^T r_{\tau}(\hat{y}_{\tau}; \hat{Y}_{1 \ldots \tau
- 1})$ as a target to $\hat{Q}(\hat{y}_t; \hat{Y}_{1 \ldots t - 1})$, and use the
critic parameters $\phi$ to minimize the square error between these two values.
However, like with REINFORCE, using such a target yields to
very high variance which quickly grows with the number
of steps $T$.
We use a temporal difference (TD) method for policy
evaluation~\citep{sutton1988learning}. Namely, we use the right-hand side 
$q_t=r_t(\hat{y}_t;\hat{Y}_{1 \ldots t - 1}) + \sum_{a \in A}
p(a|\hat{Y}_{1 \ldots t}) \hat{Q}(a;\hat{Y}_{1 \ldots t})$ of the Bellman equation as the
target for the left-hand $\hat{Q}(\hat{y}_t; \hat{Y}_{1 \ldots t - 1})$.


\paragraph{Applying deep RL techniques}

It has been shown in the RL literature that if $\hat{Q}$ is non-linear (like 
in our case), the TD policy evaluation might diverge \citep{tsitsiklis1997analysis}.
Previous work has shown that this problem can be alleviated by using an additional
\emph{target network}
$\hat{Q}'$ to compute $q_t$, which is updated less often and/or more slowly than $\hat{Q}$. 
Similarly to \citep{lillicrap2015continuous}, we update the parameters $\phi'$
of the target critic by linearly interpolating them with the parameters of the trained
one.
Attempts to remove the target network by propagating the gradient through $q_t$ 
resulted in a lower square error 
$(\hat{Q}(\hat{y}_t;  \hat{Y}_{1 \ldots T}) - q_t)^2$, but the resulting $\hat{Q}$ 
values proved very unreliable as training signals for the actor.

The fact that both actor and critic use
outputs of each other for training creates a potentially
dangerous feedback loop. To address this, we
sample predictions from a \textit{delayed actor} \citep{lillicrap2015continuous},
whose weights are slowly updated to follow the actor that
is actually trained.

\paragraph{Dealing with large action spaces}

One of the challenges of our work is that the action space is very large (as is typically the case in
NLP tasks with large vocabularies). This can be alleviated by putting constraints on
the critic values for actions that are rarely sampled. We found experimentally
that shrinking the values of these rare actions is
 necessary for the algorithm to converge. Specifically, we add a term $C_t$ for every step $t$ to the
critic's optimization objective which drives all value predictions of the critic
closer to their mean: 
\begin{equation} 
    C_t = \sum_{a}\left( 
        \hat{Q}(a; \hat{Y}_{1\ldots t-1}) 
        - \frac{1}{|\mathcal{A}|}\sum_{b}\hat{Q}(b; \hat{Y}_{1 \ldots t-1}) 
    \right)^2 \label{eq:variance_penalty}
\end{equation} This corresponds to
penalizing the \emph{variance} of the outputs of the critic. 
Without this penalty the values of rare actions can be severely overestimated,
which biases the gradient estimates and can cause divergence. 
A similar trick was used in the context of learning 
simple algorithms with Q-learning \citep{zaremba2015learning}.

\paragraph{Reward shaping}
While we are ultimately interested in the maximization of the score of a complete
prediction, simply awarding this score at the last 
step provides a very sparse training signal for the critic. For this reason we
use potential-based reward shaping with potentials $\Phi(\hat{Y}_{1 \ldots t})
= R(\hat{Y}_{1 \ldots t})$ for incomplete sequences and $\Phi(\hat{Y}) = 0$ for
complete ones \citep{ng1999policy}. Namely, for a predicted sequence $\hat{Y}$
we compute score values for all prefixes to obtain the sequence of scores
$(R(\hat{Y}_{1\ldots 1}),R(\hat{Y}_{1\ldots 2}),\ldots,R(\hat{Y}_{1\ldots
T}))$.  The difference between the consecutive pairs of scores is then used
as the reward at each step: $r_t(\hat{y}_t;\hat{Y}_{1\ldots
t-1})=R(\hat{Y}_{1\ldots t})-R(\hat{Y}_{1\ldots t-1})$. Using the shaped reward $r_t$ 
instead of awarding the whole score $R$ at the last step does not change the 
optimal policy
\citep{ng1999policy}.
\paragraph{Putting it all together}

Algorithm \ref{algo:algo1} describes the proposed method in detail.
We consider adding the weighted
log-likelihood gradient to the actor's gradient estimate. This is in line with
the prior work by \citep{ranzato2015sequence} and \citep{shen2015minimum}. It is
also motivated by our preliminary experiments that showed that using the
actor-critic estimate alone can lead to an early determinization of the policy
and vanishing gradients (also discussed in Section \ref{sec:discuss}).
Starting training with a randomly initialized actor and critic would be
problematic,
 because neither the actor nor the critic would provide adequate
training signals for one another.
 The actor would sample completely random
predictions that receive very little reward,
 thus providing a very weak
training signal for the critic.
 A random critic would be similarly useless
for training the actor. Motivated by these considerations, 
 we pre-train the
actor using standard log-likelihood training. Furthermore, we
 pre-train the
critic by feeding it samples from the pre-trained actor, while the actor's
parameters are frozen. The complete training procedure including pre-training
is described by Algorithm \ref{algo:algo2}.

\section{Related Work}

In other recent RL-inspired work on sequence prediction, \citet{ranzato2015sequence}
trained a translation model 
by gradually transitioning from maximum likelihood learning into
optimizing BLEU or ROUGE scores using the REINFORCE algorithm.
However, REINFORCE is known to have very high variance
and does not exploit the availability of the ground-truth like the critic
network does. The approach also relies on a curriculum learning scheme.
Standard value-based RL algorithms like SARSA and OLPOMDP have also been applied to structured
prediction \citep{maes2009structured}. Again, these systems do not
use the ground-truth for value prediction.

Imitation learning has also been
applied to structured prediction \citep{vlachos2012investigation}.
Methods of this type include the \textsc{Searn} \citep{daume2009search}
and \textsc{DAgger} \citep{ross2010reduction} algorithms.
These methods rely on an 
\emph{expert policy} to provide action sequences that the policy
learns to imitate. Unfortunately, it's not always easy or even
possible to construct an expert policy for a task-specific score. 
In our approach, the critic plays a role that is similar to the
expert policy, but is learned without requiring
prior knowledge about the task-specific score.
The recently proposed `scheduled sampling'
\citep{bengio2015scheduled} can also be seen as imitation learning.
In this method, ground-truth tokens are occasionally replaced by
samples from the model itself during training. A limitation
is that the token $k$ for the ground-truth answer is used as the target at 
step $k$, which might not always be the optimal strategy.

There are also approaches that aim to approximate the gradient of the
expected score. One such approach is
`Direct Loss Minimization' \citep{hazan2010} in which the
inference procedure is adapted to take both the model likelihood and
task-specific score into account.
Another popular approach is to replace the domain over which the task score expectation is
defined with a small subset of it, as is done in Minimum (Bayes) Risk
Training \citep{goel2000minimum,shen2015minimum,och2003minimum}.
This small subset is typically an $n$-best list or a sample (like in REINFORCE) that may or may not include
the ground-truth as well.
None of these methods provide intermediate targets for the actor during
training, and \citet{shen2015minimum} report that as many as 100 samples
were required for the best results.

Another recently proposed method is to optimize a global sequence
cost with respect to the selection and pruning behavior of the beam search procedure itself
\citep{wiseman2016sequence}. This method follows the more general strategy
called `learning as search optimization' \citep{daume2005learning}. This is an
interesting alternative to our approach; however, it is designed
specifically for the precise inference procedure involved.

\section{Experiments}

To validate our approach, we performed two sets of experiments \footnote{
The source code is available at \url{https://github.com/rizar/actor-critic-public}}.
First, we trained
the proposed model to recover strings of natural text from their corrupted versions.
Specifically, we consider each character in a natural language corpus
and with some probability replace it with a random
character. We call this synthetic task
 \textit{spelling correction}.
A desirable property of this synthetic task
is that data is essentially infinite and overfitting is no concern. 
Our second series of experiments is done on the task of automatic machine translation
using different models and datasets.

In addition to maximum likelihood and actor-critic training we implemented two versions of
the REINFORCE gradient estimator. In the first version, we use a linear baseline network that takes
the actor states as input, exactly as in \citep{ranzato2015sequence}. 
We also propose a novel extension of REINFORCE that leverages the extra information available
in the ground-truth output $Y$. Specifically, we use the $\hat{Q}$ estimates produced by
the critic network as the baseline for the REINFORCE algorithm. The motivation behind this approach
is that using the ground-truth output should produce a better baseline that lowers
the variance of REINFORCE, resulting in higher task-specific scores. We refer
to this method as REINFORCE-critic. 
\subsection{Spelling Correction} 

We use text from the One Billion Word dataset for the spelling correction
task \citep{chelba2013one}, which has pre-defined training and testing sets.
The training data was abundant, and we never used any example twice.  We
evaluate trained models on a section of the test data that comprises 6075
sentences. To speed up experiments, we clipped all sentences to the first 10
or 30 characters.

For the spelling correction actor network, we use an RNN with 100 Gated
Recurrent Units (GRU) and a bidirectional GRU network for the encoder. We
use the same attention mechanism as proposed in \citep{bahdanau2015neural},
which effectively makes our actor network a smaller version of the model used
in that work. For the critic network, we employed a model
with the same architecture as the actor. 

We use character error rate (CER) to measure performance on the spelling
task, which we define as the ratio between the total of Levenshtein distances
between predictions and ground-truth outputs and the total length of the
ground-truth outputs. This is a corpus-level metric for which a lower value 
is better. We use it as the return by negating
per-sentence ratios. At the evaluation time greedy search is used to extract
predictions from the model.

\begin{wrapfigure}{r}{0.5\textwidth}
    \centering
    \includegraphics[scale=0.5]{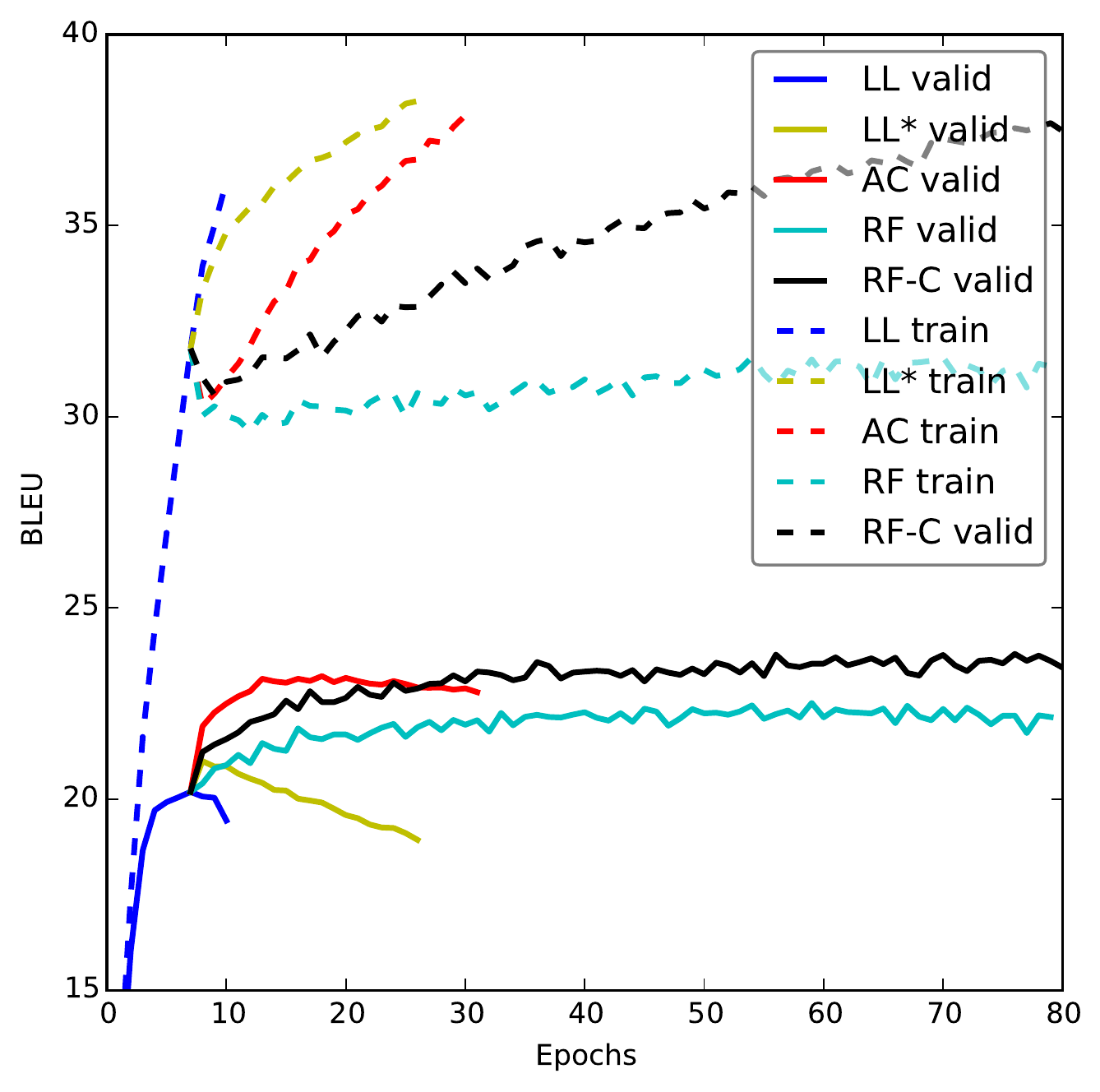}
    \caption{Progress of log-likelihood (LL), REINFORCE (RF) and actor-critic (AC) training
        in terms of BLEU score on the training (train) and
        validation (valid) datasets. LL* stands for the annealing phase of log-likelihood training. 
        The curves start from the epoch of log-likelihood
        pretraining from which the parameters were initialized. 
    }
    \label{fig:progress}       
\end{wrapfigure}    

We use the ADAM optimizer~\citep{kingma2015method} to train all the networks
with the parameters recommended in the original paper, with the exception of
the scale parameter $\alpha$. The latter is first set to $10^{-3}$ and then
annealed to $10^{-4}$ for log-likelihood training. For the pre-training
stage of the actor-critic, we use $\alpha=10^{-3}$ and decrease it to
$10^{-4}$ for the joint actor-critic training. We pretrain the actor until
its score on the development set stops improving. We pretrain the critic until
its TD error stabilizes\footnote{A typical behaviour for TD error was to grow
at first and then start decreasing slowly.  We found that stopping pretraining
shortly after TD error stops growing leads to good results.}.  We used $M=1$
sample for both actor-critic and REINFORCE. For exact hyperparameter settings
we refer the reader to Appendix \ref{sec:hyperparameters}.

We start REINFORCE training from a pretrained actor, but we do
not use the curriculum learning employed in MIXER. The critic is trained in the
same way for both REINFORCE and actor-critic,
including the pretraining stage.
We report results obtained with the reward shaping described in Section
\ref{sec:actor-critic}, as we
found that it slightly improves REINFORCE performance.

Table \ref{tab:spelling} presents our results on the spelling correction task.
We observe an improvement in CER over log-likelihood training for all four settings considered.
Without simultaneous log-likelihood training,
actor-critic training results in a better CER than REINFORCE-critic 
in three out of four settings.
In the fourth case, actor-critic and REINFORCE-critic have similar
performance. Adding the log-likelihood gradient with a cofficient
$\lambda_{LL}=0.1$ helps both of the methods,
but actor-critic still retains a margin of improvement over REINFORCE-critic.
\subsection{Machine Translation}

For our first translation experiment, we use data from the German-English
machine translation track of the IWSLT 2014 evaluation campaign
\citep{cettolo2014report}, as used in \citet{ranzato2015sequence}, and closely follow the pre-processing described in
that work. The training data comprises about 153,000 German-English
sentence pairs. In addition we considered a larger WMT14 English-French dataset 
\cite{cho2014learning} with more than 12 million examples. For further information
about the data we refer the reader to Appendix \ref{sec:data}.

The return is defined as a smoothed and rescaled version of the BLEU score. Specifically,
we start all n-gram counts from 1 instead of 0, and multiply the resulting score
by the length of the ground-truth translation. Smoothing is a common practice when sentence-level
BLEU score is considered, and it has been used to apply REINFORCE in similar settings
\citep{ranzato2015sequence}.

\paragraph{IWSLT 2014 with a convolutional encoder}

In our first experiment we use a convolutional encoder in the actor to make our results more comparable with \citet{ranzato2015sequence}.
For the same reason, we use 256 hidden units in the  networks.
For the critic, we replaced the convolutional network with a bidirectional GRU network.
For training this model we mostly used the same hyperparameter values as in the spelling 
correction experiments, with a few differences highlighted in Appendix \ref{sec:hyperparameters}.
For decoding we used greedy search and beam search with a beam size of 10. We
found that penalizing
 candidate sentences that are too short was required to
obtain the best results. Similarly to \citep{hannun2014first},
 we subtracted
$\rho T$ from the negative log-likelihood of each candidate
 sentence, where
$T$ is the candidate's length, and $\rho$ is a hyperparameter
 tuned on the
validation set.

\begin{table}
    \centering
    \caption{Character error rate of different methods on the spelling correction task.
        In the table $L$ is the length of input strings, $\eta$ 
        is the probability of replacing a character with a random one.
        LL stands for the log-likelihood training, AC and RF-C and for the actor-critic and the
        REINFORCE-critic respectively, AC+LL and RF-C+LL for the combinations of AC and RF-C with LL.}
    \begin{tabular}{l | c | c | c | c | c}
        \multirow{2}{*}{Setup} & \multicolumn{3}{|c}{Character Error Rate} \\
                               & LL & AC & RF-C & AC+LL & RF-C+LL \\
        \hline\hline
        $L=10, \eta=0.3$  & 17.81 & 17.24 & 17.82 & \textbf{16.65} & 16.97 \\
        $L=30, \eta=0.3$  & 18.4  & 17.31 & 18.16 & \textbf{17.1} & 17.47 \\
        $L=10, \eta=0.5$  & 38.12 & 35.89 & 35.84 & \textbf{34.6} & 35 \\
        $L=30, \eta=0.5$  & 40.87 & 37.0 & 37.6 & \textbf{36.36} & 36.6 
    \end{tabular}
    \label{tab:spelling}
\end{table}

\begin{table}
    \centering
    \caption{Our IWSLT 2014 machine translation results with a convolutional encoder compared to the previous work by Ranzato et al.
    Please see \ref{tab:spelling} for an explanation of abbreviations. The asterisk
    identifies results from \citep{ranzato2015sequence}. 
    The numbers reported with $\leq$ were approximately read from
    Figure 6 of \citep{ranzato2015sequence} }
    \begin{tabular}{l | c | c | c | c | c | c}
        \multirow{2}{*}{Decoding method} & \multicolumn{6}{|c}{Model} \\
                                         & LL* & MIXER* & LL & RF & RF-C & AC \\
        \hline\hline
        greedy search & 17.74       & 20.73       & 19.33 & 20.92 & \textbf{22.24} & 21.66 \\
        beam search   & $\leq$ 20.3 & $\leq$ 21.9 & 21.46 & 21.35 & \textbf{22.58} & 22.45 
    \end{tabular}
    \label{tab:mt}
\end{table}

\begin{table}
    \centering
    \caption{
        Our IWSLT 2014 machine translation results with a bidirectional recurrent encoder compared to the previous work.
        Please see Table \ref{tab:spelling} for an explanation of abbreviations. The asterisk
        identifies results from \citep{wiseman2016sequence}. 
    }
    \begin{tabular}{l | c | c | c | c | c | c | c}
        \multirow{2}{*}{Decoding method} & \multicolumn{6}{|c}{Model} \\
                                         & LL* & BSO* & LL & RF-C & RF-C+LL & AC & AC+LL \\
        \hline\hline
        greedy search & 22.53 & 23.83 & 25.82 & 27.42 & \textbf{27.7} & 27.27 & 27.49\\
        beam search   & 23.87 & 25.48 & 27.56 & 27.75 & 28.3 & 27.75 & \textbf{28.53}
    \end{tabular}
    \label{tab:mt2}
\end{table}

\begin{table}
    \centering
    \caption{
        Our WMT 14 machine translation results compared to the previous work.
        Please see Table \ref{tab:spelling} for an explanation of
        abbreviations. The apostrophy and the asterisk identify results from
        \citep{bahdanau2015neural} and \citep{shen2015minimum} respectively.
    }
    \begin{tabular}{l | c | c | c | c | c | c}
        \multirow{2}{*}{Decoding method} & \multicolumn{6}{|c}{Model} \\
                                         & LL' & LL* & MRT * & LL & AC+LL & RF-C+LL \\
        \hline\hline
        greedy search & n/a & n/a & n/a & 29.33 & \textbf{30.85} & 29.83\\
        beam search   & 28.45 & 29.88 & \textbf{31.3} & 30.71 & 31.13 & 30.37
    \end{tabular}
    \label{tab:wmt}
\end{table}

The results are summarized in Table \ref{tab:mt}. We report
a significant improvement of $2.3$ BLEU points over the log-likelihood baseline
when greedy search is used for decoding. Surprisingly, the best performing method
is REINFORCE with critic, with an additional $0.6$ BLEU point advantage
over the actor-critic. When beam-search is used, the ranking of the compared approaches
is the same, but the margin between the proposed methods and log-likelihood training
becomes smaller.  The final performances of the actor-critic and the
REINFORCE-critic with greedy search are also $0.7$ and $1.3$ BLEU points
respectively better than what \citet{ranzato2015sequence} report for their
MIXER approach.  This comparison should be treated with caution,
because our log-likelihood baseline is $1.6$ BLEU points stronger
than its equivalent from \citep{ranzato2015sequence}. The performance
of REINFORCE with a simple baseline matches the score reported for MIXER in
\citet{ranzato2015sequence}.

To better understand the IWSLT 2014 results we provide
the learning curves for the considered approaches in Figure
\ref{fig:progress}. We can clearly see that the training methods that use
generated predictions have a strong regularization effect --- that is, better
progress on the validation set in exchange for slower or negative progress on
the training set. The effect is stronger for both REINFORCE varieties, 
especially for the one without a critic. The actor-critic training does a 
much better job of fitting the training set than REINFORCE and is the only
method except log-likelihood that shows a clear overfitting, which is a healthy 
behaviour for such a small dataset.

In addition, we performed an ablation study.  We found that using a target network was crucial;  while
the joint actor-critic training was still progressing with $\gamma_{\theta}=0.1$, 
with $\gamma_{\theta}=1.0$ it did not work at all. Similarly
important was the value penalty described in Equation
\eqref{eq:variance_penalty}. We found that good values of the $\lambda$ coefficient
were in the range $[10^{-3}, 10^{-6}]$.  Other
techniques, such as reward shaping and a delayed actor, brought moderate
performance gains. We refer the reader to Appendix \ref{sec:hyperparameters} for more details.

\paragraph{IWSLT 2014 with a bidirectional GRU encoder}
In order to compare our results with those reported by
\citet{wiseman2016sequence} we repeated our IWSLT 2014 investigation with a
different encoder, a bidirectional RNN with 256 GRU units. In this round of
experiments we also tried to used combined training objectives in the same way as in
our spelling correction experiments. The results are summarized in Table \ref{tab:mt2}.
One can see that the actor-critic training, especially its AC+LL version, yields significant improvements (1.7 with greedy search and 1.0 with beam search) upon the pure log-likelihood training, which are comparable to those brought by Beam Search Optimization (BSO), even though our log-likelihood baseline is much stronger. In this round of experiments actor-critic and REINFORCE-critic performed on par.

\paragraph{WMT 14}
Finally we report our results on a very popular
large WMT14 
English-French dataset \citep{cho2014learning} in Table \ref{tab:wmt}. Our
model closely follows the
achitecture from \citep{bahdanau2015neural}, however we achieved a higher
baseline performance by annealing the learning rate $\alpha$ and penalizing 
output sequences that were too short during beam search. The actor-critic training brings a
significant 1.5 BLEU improvement with greedy search and a noticeable 0.4 BLEU
improvement with beam search. In previous work \cite{shen2015minimum}
report a higher improvement of 1.4 BLEU with beam search, however they use 100
samples for each training example, whereas we use just one. We note that in this experiment, which is perhaps the most realistic settings, the actor-critic enjoys a significant advantage over the REINFORCE-critic.

\section{Discussion}
\label{sec:discuss}

We proposed an actor-critic approach to sequence prediction. Our method takes
the task objective into account during training and uses the ground-truth output to aid
the critic in its prediction of intermediate targets for the actor. We showed that our method leads to
significant improvements over maximum likelihood training on both a synthetic task
and a machine translation benchmark. Compared to REINFORCE training on machine translation,
actor-critic fits the training data much faster, although in some of our
experiments we were able to significantly reduce the gap in the training speed and
achieve a better test error using our critic network as the baseline for
REINFORCE.


One interesting observation we made from the machine translation results is
that the training methods that use generated predictions have a strong
regularization effect. Our understanding is that conditioning on the sampled
outputs effectively increases the diversity of training data. This phenomenon
makes it harder to judge whether the actor-critic training meets our
expectations, because a noisier gradient estimate yielded a better test set
performance. We argue that the spelling correction results obtained on
a virtually infinite dataset in conjuction with better machine translation
performance on the large WMT 14 dataset provide convincing evidence that the
actor-training can be effective. In future work we will consider larger
machine translation datasets.

We ran into several optimization issues.
The critic would sometimes assign very high values to
actions with a very low probability according to the actor. We were able to
resolve this by penalizing the critic's variance. Additionally, 
the actor would sometimes have trouble to adapt to the demands of
the critic. We noticed that the action distribution tends to saturate
and become deterministic, causing the gradient to vanish. We found that
combining an RL training objective with log-likelihood can help, but in general
we think this issue deserves further investigation. For example, one can look
for suitable training criteria that have a well-behaved gradient even when the
policy has little or no stochasticity.

In a concurrent work \citet{wu2016google} show that a version of REINFORCE with
the baseline computed using multiple samples can improve performance of a very
strong machine translation system. This result, and our REINFORCE-critic
experiments, suggest that often the variance of REINFORCE can be reduced
enough to make its application practical. That said, we would like to emphasize
that this paper attacks the problem of gradient estimation from a very
different angle as it aims for low-variance but potentially high-bias
estimates. The idea of using the ground-truth output that we proposed is an
absolutely necessary first step in this direction. Future work could focus on
further reducing the bias of the actor-critic estimate, for example, by using a
multi-sample training criterion for the critic.


\subsubsection*{Acknowledgments}
We thank the developers of Theano~\citep{team2016theano} and Blocks~\citep{blocksfuel} for
their great work. We thank NSERC, Compute Canada, Calcul Queb\'ec, Canada Research
Chairs, CIFAR, CHISTERA project M2CR (PCIN-2015-226) and Samsung Institute of Advanced Techonology for their financial support.

\bibliography{references}
\bibliographystyle{iclr2017_conference}

\newpage
\appendix
\section{Hyperparameters}
\label{sec:hyperparameters}

For machine translation experiments the variance penalty coefficient $\lambda$
was set to $10^{-4}$, and the delay coefficients $\gamma_{\theta}$ and
$\gamma_{\phi}$ were both set to $10^{-4}$. For REINFORCE with the critic we did not use 
a delayed actor, i.e. $\gamma_{\theta}$ was set to 1. For the spelling correction
task we used the same $\gamma_{\theta}$ and $\gamma_{\phi}$ but a different 
$\lambda=10^{-3}$. When we used a combined training criterion, the weight of the log-likelihood gradient $\lambda_{LL}$ was always 0.1. All initial weights were sampled from a centered uniform distribution with
width $0.1$.

In some of our experiments we provided the actor states as additional inputs to
the critic.  Specifically, we did so in our spelling correction experiments and
in our WMT 14 machine translation study. All the other results were obtained
without this technique.

For decoding with beam search we substracted the length of a candidate
times $\rho$ from the log-likelihood cost. The exact value of $\rho$ 
was selected on the validation set and was equal to $0.8$
for models trained by log-likelihood and REINFORCE and to $1.0$
for models trained by actor-critic and REINFORCE-critic.

For some of the hyperparameters we performed an ablation study. The results are reported in Table \ref{tab:ablation}.

\begin{table}
    \caption{Results of an ablation study. We tried varying the actor update speed $\gamma_{\theta}$, 
             the critic update speed $\gamma_{\phi}$, the value penalty coefficient $\lambda$, 
             whether or not reward shaping is used, whether or not temporal difference (TD) learning is used
             for the critic. Reported are the best training and validation BLEU score obtained in the course
             of the first 10 training epochs. Some of the validation scores would still improve with longer training.
             Greedy search was used for decoding.}
    \centering
    \begin{tabular}{c  c  c  c  c  c c}
        $\gamma_{\theta}$ & $\gamma_{\phi}$ & $\lambda$ & reward shaping & TD & train BLEU & valid BLEU \\
        \hline
        \multicolumn{7}{c}{baseline} \\
        0.001 & 0.001 & $10^{-3}$ & yes & yes & 33.73 & 23.16 \\
        \hline
        \multicolumn{7}{c}{with different $\gamma_{\phi}$} \\
        0.001 & 0.01 & $10^{-3}$ & yes & yes & 33.52 & 23.03 \\
        0.001 & 0.1 & $10^{-3}$ & yes & yes & 32.63 & 22.80 \\
        0.001 & 1 & $10^{-3}$ & yes & yes & 9.59 & 8.14 \\
        \hline
        \multicolumn{7}{c}{with different $\gamma_{\theta}$} \\
        1 & 0.001 & $10^{-3}$ & yes & yes & 32.9 & 22.88 \\
        \hline
        \multicolumn{7}{c}{without reward shaping} \\
        0.001 & 0.001 & $10^{-3}$ & no & yes & 32.74 & 22.61 \\
        \hline
        \multicolumn{7}{c}{without temporal difference learning} \\
        0.001 & 0.001 & $10^{-3}$ & yes & no & 23.2 & 16.36 \\
        \hline
        \multicolumn{7}{c}{with different $\lambda$} \\
        0.001 & 0.001 & $3 * 10^{-3}$ & yes & yes & 32.4 & 22.48 \\
        0.001 & 0.001 & $10^{-4}$ & yes & yes & 34.10 & 23.15 \\
        0.001 & 0.001 & $10^{-6}$ & yes & yes & 35.00 & 23.10 \\
        0.001 & 0.001 & $10^{-8}$ & yes & yes & 33.6 & 22.72 \\
        0.001 & 0.001 & $0$ & yes & yes & 27.41 & 20.55 \\
    \end{tabular}
    \label{tab:ablation}
\end{table}

\section{Data}
\label{sec:data}
For the IWSLT 2014 data the sizes of validation and tests set were 6,969 and 6,750, 
respectively. We limited the number of words in the English and German
vocabularies to the 22,822 and 32,009 most frequent words, respectively, and
replaced all other words with a special token. The maximum sentence length
in our dataset was 50. For WMT14 we used vocabularies of 30,000 words for both English and French, and the maximum sentence length was also 50.

\section{Generated Q-values}

In Table \ref{tab:qvalues} we provide an example of value predictions
that the critic outputs for candidate next words. One can see that the critic has
indeed learnt to assign larger values for the appropriate next words.
While the critic does not always produce sensible estimates and can often predict
a high return for irrelevant rare words, this is greatly reduced using the variance
penalty term from Equation \eqref{eq:variance_penalty}.

\begin{figure}[h]
\centering
\label{tab:qvalues}
\caption{The best 3 words according to the critic at intermediate
        steps of generating a translation. The numbers in parentheses
        are the value predictions $\hat{Q}$. The German original is 
        ``\"{u}ber eine davon will ich hier erz\"{a}hlen .''
        The reference translation is 
        ``and there's one I want to talk about''.}

\begin{tabular}{c|c}
    \textrm{Word} & \textrm{Words with largest $\hat{Q}$} \\
    \hline
    one &
    and(6.623) there(6.200) but(5.967) \\
    of &
    that(6.197) one(5.668) \&apos;s(5.467) \\
    them &
    that(5.408) one(5.118) i(5.002) \\
    i &
    that(4.796) i(4.629) ,(4.139) \\
    want &
    want(5.008) i(4.160) \&apos;t(3.361) \\
    to &
    to(4.729) want(3.497) going(3.396) \\
    tell &
    talk(3.717) you(2.407) to(2.133) \\
    you &
    about(1.209) that(0.989) talk(0.924) \\
    about &
    about(0.706) .(0.660) right(0.653) \\
    here &
    .(0.498) ?(0.291) --(0.285) \\
    . &
    .(0.195) there(0.175) know(0.087) \\
     $\emptyset$  &
    .(0.168)  $\emptyset$ (-0.093) ?(-0.173) \\
\end{tabular}
\end{figure}

\newpage

\section{Proof of Equation \eqref{eq:ac}}

\begin{align*}
    \frac{dV}{d\theta} = 
    \frac{d}{d\theta} \Exp_{\hat{Y} \sim p(\hat{Y})} R(\hat{Y}) =
    \sum\limits_{\hat{Y}}
    \frac{d}{d\theta}
    \left[
    p(\hat{y}_1) p(\hat{y}_2|\hat{y}_1) 
    \ldots
    p(\hat{y}_T|\hat{y}_1 \ldots \hat{y}_{T-1})
    \right]
    R(\hat{Y}) = \notag\\
    \sum\limits_{t=1}^T
    \sum\limits_{\hat{Y}}
    p(\hat{Y}_{1 \ldots t-1})
    \frac{dp(\hat{y}_t|\hat{Y}_{1 \ldots t - 1})}{d\theta}
    p(\hat{Y}_{t+1 .. T}|\hat{Y}_{1 \ldots t})
    R(\hat{Y}) = \notag\\
    \begin{aligned}
        & \sum\limits_{t=1}^T
          \sum\limits_{\hat{Y}_{1 \ldots t}}
          p(\hat{Y}_{1 \ldots t-1})
          \frac{dp(\hat{y}_t|\hat{Y}_{1 \ldots t - 1})}{d\theta}
        &  \sum\limits_{\hat{Y}_{t + 1 \ldots T}}
           p(\hat{Y}_{t+1 .. T}|\hat{Y}_{1 \ldots t})
           \sum\limits_{\tau=1}^T 
           r_{\tau}(\hat{y}_{\tau}; \hat{Y}_{1 \ldots \tau - 1})
    \end{aligned} 
    =\notag\\
    \begin{aligned}
        & \sum\limits_{t=1}^T
          \sum\limits_{\hat{Y}_{1 \ldots t}}
          p(\hat{Y}_{1 \ldots t-1})
          \frac{dp(\hat{y}_t|\hat{Y}_{1 \ldots t - 1})}{d\theta} \\
        &  \left[
          r_t(\hat{y}_t; \hat{Y}_{1 \ldots t - 1}) +
          \sum\limits_{\hat{Y}_{t + 1 \ldots T}}
           p(\hat{Y}_{t+1 .. T}|\hat{Y}_{1 \ldots t})
           \sum\limits_{\tau=t+1}^T 
           r_{\tau}(\hat{y}_{\tau}; \hat{Y}_{1 \ldots \tau - 1})
           \right]
    \end{aligned} 
    = \notag\\
    \sum\limits_{t=1}^T
    \Exp_{\hat{Y}_{1 \ldots t - 1} \sim p(\hat{Y}_{1 \ldots t - 1})}
    \sum\limits_{a \in A}
    \frac{dp(a|\hat{Y}_{1 \ldots t - 1})}{d\theta}
    Q(a;\hat{Y}_{1 \ldots t - 1}) 
    = \notag\\
    \Exp_{\hat{Y} \sim p(\hat{Y})} 
    \sum\limits_{t=1}^T
    \sum\limits_{a \in \mathcal{A}}
    \frac{dp(a|\hat{Y}_{1 \ldots t - 1})}{d\theta}
    Q(a;\hat{Y}_{1 \ldots t - 1})
\end{align*}

\end{document}